# Talking to oneself in CMC: a study of self replies in Wikipedia talk pages

## Ludovic Tanguy[1], Céline Poudat[2], Lydia-Mai Ho-Dac[1]

1: CLLE - CNRS & University of Toulouse, France
2: BCL - CNRS & University of Nice Côte d'Azur, France

Email: ludovic.tanguy@univ-tlse2.fr, celine.poudat@univ-cotedazur.fr, lydia-mai.do-dac@univ-tlse2.fr

**Abstract**
This study proposes a qualitative analysis of self replies in Wikipedia talk pages, more precisely when the first two messages of a discussion are written by the same user. This specific pattern occurs in more than 10% of threads with two messages or more and can be explained by a number of reasons. After a first examination of the lexical specificities of second messages, we propose a seven categories typology and use it to annotate two reference samples (English and French) of 100 threads each. Finally, we analyse and compare the performance of human annotators (who reach a reasonable global efficiency) and instruction-tuned LLMs (which encounter important difficulties with several categories).

## 1. Introduction

Wikipedia Talk pages have been extensively studied as they provide a unique means to examine the dynamics of interaction between users in their collaborative efforts to contribute to the online encyclopaedia (Laniado et al. 2011, Gomez et al. 2011, Lüngen & Herzberg 2019, Schneider et al. 2010, Kopf 2022).

In this study, we propose to focus on a very specific and understudied practice in Wikipedia talks and written CMC studies in general: monologues in interactional settings i.e. situations in which users persist in posting even when no one else is intervening. Such situations are close to the phenomenon of "no response", which would not be uncommon in other written CMC genres such as forums for instance - Beaudouin and Velkovska were already observing in 1999 that 41% of the posts in a forum had not received any response within one month. Monologues in interactional settings are quite different from other situations which have been described so far, such as teacher monologues in the classroom or in vlogs for instance (e.g. Frobenius 2011). Indeed, these situations are intrinsically monological, which is not the case in Wikipedia talks.

To explore this phenomenon, we decided to start from the beginning, and concentrate on monologue inceptions. To achieve this, we analyse thread onsets in which users technically reply to themselves in a thread they have initiated. We focus on the reasons why a user would post a second message to himself/herself. We first provide an overview of this phenomenon in Wikipedia and examine the specific lexicon found in the posts. We then propose a typology of the seven main reasons we have identified for a self reply. We asked human coders to apply this typology on a sample of 100 threads. We analyse the results and finally present a first try at using Large Language Models to annotate the large amount of data available and discuss the overall difficulty of the annotation task.

## 2. Corpus and phenomenon overview

We base our study on the *EFGCorpus*, a comparable corpus composed of all the talk pages extracted from the August 2019 dump of the English, French and German versions of Wikipedia (Ho-Dac, to appear).
All the talk pages included in the EFGCorpus are encoded according to the TEI CMC-core schema (Beißwenger and Lüngen 2020). For each language, we selected all the threads in which there are neither unsigned and undated messages nor messages signed by a bot.

As Wikipedia talk pages are written the same device as the article pages, and therefore allow for a loose structure, we considered the linear order of messages as they appear in each thread. Table 1 gives an overview of the corpus and details concerning consecutive posts by the same author. All three languages show similar tendencies, although the quantities vary according to the size of the Wikipedia communities.

| Feature | English | French | German |
|---|---|---|---|
| **Number of threads** | 3 385 583 | 302 475 | 1 485 648 |
| **Number of posts** | 8 873 620 | 769 880 | 3 967 726 |
| **Threads with 2 posts or more** | 1 688 939 | 140 904 | 784 605 |
| **Threads containing two consecutive posts by the same author** | 406 292 (24,1%) | 38 706 (27,5%) | 179 871 (22,9%) |
| **Threads starting with two consecutive posts by the same author** | 201 280 (11,9%) | 19 947 (14,2%) | 82 629 (10,5%) |
| **Single-author threads with 2 posts or more** | 115 813 (6,9%) | 12 019 (8,5 %) | 48 413 (6,2%) |

Table 1: Quantitative data overview

Half of the threads in the corpus contain only one post and thus are not conversations, whether because they have failed (no one answered the user who left the discussion as it is) or do not require a response (e.g. the post contains simple information). Among the subset of threads which contain at least two messages, about 20-25% contain at least one pair of consecutive posts by the same author. We decided to focus on an even more specific subset by selecting the threads which begin with two messages by the same author. This phenomenon is clearly very frequent as these represent more than 10% of the threads with two or more messages. Finally, we can see that 6-8% of all threads in our corpus are purely monological, which is significant.

Figure 1 shows an example thread in English in which the author of the first message (*198.6.46.11*) replies in the second message[1].

---
[1] Unregistered Wikipedia users are identified by their IP address. Therefore it is possible that we missed a number of self replies, as a user's IP can change between two posts. On the other hand, false positives are very highly improbable.

| | |
|---|---|
| Rose McGowan? | |
| In the notable people from Eugene, McGowan is listed but her Wiki entry doesn't state anything about Eugene. Anyone know the story on this? 198.6.46.11 (talk) 17:29, 2 September 2008 (UTC) | |
|    Alright then, removed 198.6.46.11 (talk) 18:47, 7 October 2008 (UTC) | |

Figure 1: Example single author thread
https://en.wikipedia.org/wiki/Talk:Eugene,_Oregon/Archive_2#Rose_McGowan?

In this example, the user opened a thread in order to highlight a lack of information or coherence and replied one month later to his first message using an interactional marker (*Alright then*) in order to report that he solved the problem by deleting the aforementioned section in the article. We estimated that focusing on self reply at the beginning of a thread is a more direct approach to the phenomenon, and also easier to investigate, as it does not require to follow a sometimes lengthy discussion.

We will now concentrate on our major research question: what are the reasons why the user posts a second message as a reply to his/her first one?

## 3. Self replies in Wikipedia - particularities and motivations

Monologues have been mostly studied in speech contexts, like in classroom interactions (see for instance Davis 2007) or video blogs or vlogs (Frobenius 2011, 2014) in CMC situations. Nevertheless, they have not been studied in written CMC interactions to our knowledge.
We propose to characterise monologue onsets in Wikipedia talks. After a quick glance at the most significant lexical specificities of the second message compared to the first, we report an annotation experiment aiming at categorising the main reasons why the user has posted a second message in a thread he initiated.

### 3.1 Keyness analysis of second messages

As shown in Table 2, the user's second message is significantly distinct from the first they posted, while it explicitly refers to it. Users are completely aware they are posting a second message, as we can observe with the words *PS*, *p.s.*, *update* or *forgot*, which explicitly complement or rectify their initial post. The use of the present perfect, which has an evaluative value, is also noteworthy (*I've done / fixed / found / removed…*), as the Wikipedians report something they have done following their initial message - something they had possibly requested to be done, but which was still undone.

| Word | Keyness | Word | Keyness |
|---|---|---|---|
| 've | 1000 | ah | 244 |
| above | 1000 | mind | 227 |
| ahead | 1000 | also | 226 |
| done | 1000 | added | 221 |
| fixed | 1000 | you | 215 |
| I | 1000 | update | 207 |
| nevermind | 1000 | response | 187 |
| now | 1000 | oops | 169 |
| oh | 1000 | went | 147 |
| OK | 1000 | ! | 140 |
| okay | 1000 | update | 136 |
| PS | 308 | further | 134 |
| still | 293 | nobody | 133 |
| again | 284 | objections | 133 |
| p.s | 282 | may | 131 |
| sorry | 269 | forgot | 119 |
| found | 265 | no | 119 |
| removed | 262 | thanks | 119 |

Table 2: Main specificities[2] of the user's second message

We also note that classical interaction marks are present - although users are technically replying to themselves, the speech remains explicitly addressed, with the use of *you*, *nobody* or even *objections*. Even more strikingly, the user appears to utilise interactional markers like *OK* for instance to respond to themselves. Even in a monologue, the framework of the conversation seems to prevail.

### 3.2 A typology of self replies

We carried out a detailed examination of the content of the threads beginning with a self reply, randomly selected in the English and the French parts of the xxxCorpus. After a first exploratory stage during which two of us annotated 200 threads (100 in English, 100 in French), we established a first typology of the reasons why users reply to themselves. This typology was then finalised in a second step of curation. As a result, 7 main reasons have been identified and defined as follows:

- **Addendum:** the user complements their first message with new information, a new scope, additional arguments or some kind of clarification;
- **Self-correction**: the user has identified an error in their first message and corrects it, possibly cancelling the first message;
- **Self-answer:** the user answers the question they asked in the first message;
- **Chasing up**: having received no replies to his first message, the user asks other users for answers or reactions;
- **Action report**: the user has done something since their first message and announces it;
- **Reaction to event**: something has been done by someone else, or has happened since the first message and the user reacts to this event;
- **List**: the first two messages constitute a list of items or the beginning of a list; these items can be pieces of information, things to do, remarks, questions etc.

If the first four categories were expected and could be observed in other online discussion platforms, the last three seem more specific to Wikipedia talk pages. **Action report** is crucial in the context of collaborative working. Suggesting, requesting or validating an article edit are amongst the main reasons why users chat in talk pages, as stated in Ferschke (2014) who showed that around 60% of the messages are associated with explicit performative speech acts. As shown previously, the user who requests an action and the user who performs it may be the same. In this situation, the content of the second message may be reduced to a minimum (*Done*) or it may contain details of the action performed (as in Figure 1).
The **Reaction to event** category differs from the others in that it goes beyond the framework of mere discussion. In Wikipedia, interactions may indeed spread over multiple channels. In this category, the user writes a second message in reaction to what someone else has done in another space e.g. mostly an article edit or sometimes a message in another discussion channel. Such cases are often difficult to understand because of the lack of contextual information, as in Figure 2 where *Til Eulenspiegel* addresses his second message to someone who "*revert[ed] a valid information*", using a second-person address[3] - this may also explain the prevalence of *you* in the keyness analysis.

---
[2] The index is based on a calculation grounded on the hypergeometric distribution, using the *textometry* R package (https://cran.r-project.org/package=textometry).

> **Scholars talking about Solomon's caravan trade with Sheba** [ edit ]
>
> I have only barely scratched the surface of scholars talking about this. Some editors at RSM have taken it on themselves to say what scholarship they find acceptable. This will not be possible without a fight and a full demonstration of what they are attempting here. Til Eulenspiegel /talk/ 19:16, 1 October 2013 (UTC) [ reply ]
>
> So you are not even going to make a case on the talk page, you are just going to revert valid information pretending a "consensus"? You clearly have no idea what scholars have said on this subject. Til Eulenspiegel /talk/ 20:19, 1 October 2013 (UTC) [ reply ]

Figure 2: Example of a **Reaction to event**

https://en.wikipedia.org/wiki/Talk:Sheba#Scholars_talking_about_Solomon's_caravan_trade_with_Sheba

The **List** category is usually found in long monologues in which the users uses a thread as a logbook, a dashboard or a personal to-do list[4]. In such threads, the user just wants to keep trace of a work in progress without any intention of calling on the intervention of another user (cf. the self-commitment category in Ferschke 2014). In Figure 3, *Gurdjieff* lists all the edits he did on the 'Uruk' article.

> **edits for clarity** [ edit ]
>
> I did some edits to fix the problems with the dates and added the first citations also fixed some ambiguity in the growth section--Gurdjieff (talk) 04:12, 19 August 2008 (UTC) [ reply ]
>
> I have made many edits for clarity nothing was deleted only moved to the paragraph with the matching topic sentence. wherever I could cite a date population or land area I added this information. I also fixed the lead in sentance to meet wiki standards this article still needs alot of work for example when why and how did kullaba form? what happens to uruk after 2000bce? when was the city walled and why? ect.--Gurdjieff (talk) 00:24, 9 September 2008 (UTC) [ reply ]

Figure 3: Example of a **List**

https://en.wikipedia.org/wiki/Talk:Uruk#edits_for_clarity

A last category had to be added: **Error** is used when there is only one message, when the first two messages are not written by the same author or are unrelated (i.e. do not belong to the same thread), which can be due to various factors (syntactic anomalies, noncompliance with editing conventions…).

### 3.3 Adjudicated dataset for English and French

Once this first typology was created on the basis of a first exploratory annotation by two of us, an adjudication phase allows us to provide a consensual dataset. Table 3 shows that in both languages, the two main reasons why a user writes a second message in a thread he has opened are to complement his first message (**Addendum**) or to report an action he did (**Action report**). Half of the annotated messages could be explained by these two reasons.

---
[3] Note that this user was generically addressed as third-person (*some editors at RSM*, *they*) in the first message.
[4] Wikipedians are supposed to use dedicated "to-do" talk pages rather than the main talk pages for listing the things to do in the article, see https://en.wikipedia.org/wiki/Wikipedia:To-do_list.

The third more frequent label is the **Error** category which means that 16% for English and 11% for French of our thread do not actually begin with a self reply (due to processing errors or specific configurations).
**Self-correction**, **self-answer** or **Chasing up** are the least frequent categories.

| Categories | English | French |
|---|---|---|
| **Addendum** | 30 | 24 |
| **Action report** | 24 | 26 |
| **Reaction to event** | 8 | 14 |
| **List** | 8 | 10 |
| **Self-correction** | 4 | 11 |
| **Self-answer** | 6 | 2 |
| **Chasing up** | 4 | 2 |
| **processing error** | 16 | 11 |
| **Total** | 100 | 100 |

Table 3: Adjudicated annotation of the second message for English and French

### 3.4 Human annotation and inter-annotator agreement

We asked two students in linguistics to apply the typology to the French and English dataset and measured the inter-annotator agreements between the adjudicated and the student annotations. The students spent around 3 hours each to annotate the 100 posts with a simple task: for each thread, they were asked to focus on each first and second message independently of the rest of the talk (i.e. whether there is a third message, by whom and for what reason), attempting to identify the second message's main function. We compared their annotations with the adjudicated categories described above and obtained Cohen's Kappa scores of 0.67 for French and 0.69 for English. We considered that this validates our typology and annotation guidelines.

Table 4 gives detailed F1 scores per category. The category with the highest agreement is **Action report** (F1=0.88 FR / 0.84 EN). For the French set, the three most divergent categories are the less frequent one i.e. **Chasing up** (F1=0.33), **Self-answer** (F1=0.50) and **Self-correction** (F1=0.59). It is also the case in English for the categories **Self-correction** (F1=0.54) and **Reaction to event** (F1=0.55). It has already been clearly demonstrated that the rarer a category is, the more difficult the item is to annotate (cf. Paun et al. 2022).

| Category | Anotator 1 (French, k=0.67) | Annotator 2 (English, k = 0.69) | Mistral openorca (English, k=0.17) |
|---|---|---|---|
| **Addendum** | 0.65 | 0.71 | 0.55 |
| **Self-correction** | 0.59 | 0.54 | 0.57 |
| **Self-answer** | 0.50 | 0.67 | 0.00 |
| **Chasing up** | 0.33 | 0.80 | 0.15 |
| **Action report** | 0.88 | 0.84 | 0.39 |
| **Reaction to event** | 0.67 | 0.55 | 0.17 |
| **List** | 0.60 | 0.57 | 0.00 |
| **Macro-average F1** | 0.60 | 0.67 | 0.28 |

Table 4: Detailed F1 scores per category for two human annotators and the best LLM.

In any case, the human annotators obtained much better scores than the LLMs, which we document in the next section.

## 3.5 Classification by Large Language Models (LLM)

It is a well-known fact that recent advances in NLP technology allow for efficient and flexible systems that can annotate text data for complex phenomena (Alizadeh et al. 2023). We wanted to estimate the difficulty of automatically classifying the monological thread beginnings, which would allow us to have a larger dataset and investigate the phenomenon further.

For this experiment we selected seven generic instruction-trained Large Language Models, limiting our choice to the smaller open-source models that could be run locally on a workstation with a small GPU (up to 13 billion parameters with quantization)[5]. We prompted these LLMs with a zero-shot approach (i.e. without examples) with the following instructions:

*You are an expert linguist specialised in the study of online interactions. You will annotate online discussions from the Wikipedia talk pages where the same user replies to himself, and identify the main reason for this, using the following seven categories:*
[Description of categories as in the bulleted list in § 3.2]
*Below are the first two messages of a discussion (indicated by <MSG1> and <MSG2>). You will answer with the chosen category number for the second message, and only this number, without details nor explanation. You can decide that there is not enough data for answering and give a "NULL" answer.*

Each thread in the English adjudicated dataset was processed independently and the answers had to be manually interpreted in most cases as they rarely respected the requested format. We compared these with the adjudicated annotations: we obtained Cohen's kappa scores ranging from 0 to a low maximum of 0.165 for Mistral-openorca (Lian et al. 2023). These scores clearly indicate a low efficiency of LLMs for this task, far below what our two students could achieve. The rightmost column in Table 4 gives the F1 scores for each category for the aforementioned best LLM. This particular system performed best on **Addendum** and **Self-correction**, approaching the students' scores. **Addendum** can be seen as a generic default answer, while **Self-correction** benefits from obvious linguistic cues as discussed above. On the other end, **Self-answer** and **List** categories could not be properly identified. However, this last type is clearly out of reach for the technique we considered, as a list cannot generally be detected on the sole basis of the first two messages. Human annotators had a clear advantage as they could access the whole thread and have a global view of the recurring pattern of messages. We cannot conclude for other categories when looking at the other LLMs we experimented, as each showed a very different behaviour.

## 4. Conclusion

We have proposed a first investigation of monological thread onsets in Wikipedia talk pages. This phenomenon is clearly quite frequent and interesting, and we have proposed a first typology of the seven main reasons why a user may reply to oneself. We obtained a satisfactory first trial with human annotation. Although we have no doubts that the use of LLMs would be extremely useful to annotate on a larger scale, this still needs further investigation and experimentation, notably to stabilise the categories.

We have a number of perspectives to investigate. First, we need to extend our human annotation, and we are currently enlarging our dataset. While this extension will enable us to establish a gold standard corpus, it will also allow us to identify specific cues that could be used as a pre-annotation. We will then be able to perform a first analysis of the feature for each category (length, global pattern, time profile, favoured topics or dialogue acts etc.).

On the other hand, this step is crucial to characterise longer monologues, extending the annotation to third, fourth messages and more. One of our ultimate goals would be to characterise types of monologues (entire threads) and monologue sequences (parts of threads), initially within Wikipedia, and eventually within other CMC genres.

## References

Alizadeh, M., Kubli, M., Samei, Z., Dehghani, S., Bermeo, J. D., Korobeynikova, M., & Gilardi, F. (2023). Open-source large language models outperform crowd workers and approach ChatGPT in text-annotation tasks. *arXiv preprint arXiv:2307.02179*.

Beaudouin, V., & Velkovska, J. (1999). Constitution d'un espace de communication sur Internet (forums, pages personnelles, courrier électronique...). *Réseaux. Communication-Technologie-Société*, 17(97), 121-177.

Beißwenger, M. & Lüngen, H. (2020). CMC-core: a schema for the representation of CMC corpora in TEI. *Corpus*, 20.

Davis, J. (2007). Dialogue, monologue and soliloquy in the large lecture class. *International Journal of Teaching and Learning in Higher Education*, 19(2).

Ferschke, O. (2014). *The Quality of Content in Open Online Collaboration Platforms: Approaches to NLP-supported Information Quality Management in Wikipedia*. PhD thesis, Technische Universität Darmstadt.

Frobenius, M. (2011). « Beginning a monologue: The opening sequence of video blogs ». *Journal of Pragmatics* 43: 814-27.

Frobenius, M. (2014). Audience design in monologues : How vloggers involve their viewers. *Journal of Pragmatics*, 72, 59-72.

Gómez, V., Kappen, H. J., & Kaltenbrunner, A. (2011). Modeling the structure and evolution of discussion cascades. In *Proceedings of the 22nd ACM conference on Hypertext and hypermedia* (pp. 181-190).

Ho-Dac, L.-M. (to appear). Building a comparable corpus of online discussions in Wikipedia: the EFG WikiCorpus. In C. Poudat, H. Lüngen & L. Herzberg (Eds.), *Investigating Wikipedia: linguistic corpus building, exploration and analyses*. John Benjamins.

Kopf, S. (2022). *A Discursive Perspective on Wikipedia: More than an Encyclopaedia?* Palgrave Macmillan.

Laniado, D., Tasso, R., Volkovich, Y., & Kaltenbrunner, A. (2011). When the Wikipedians talk: Network and tree structure of Wikipedia discussion pages. In *Fifth international AAAI conference on weblogs and social media*.

Lian, W. Goodson, B., Wang, G., Pentland, E., Cook, A., Vong, C. and Teknium (2023). MistralOrca: Mistral-7B

---
[5] Mistral-Openorca 7b, Mixtral 0.2 7b, Gemma 7b, Mistral 8x7b, Llama2 7b and Llama2 13b. All models were run locally through the Ollama platform (ollama.com).

Model Instruct-tuned on Filtered OpenOrcaV1 GPT-4 Dataset. *HuggingFace repository*. https://huggingface.co/Open-Orca/Mistral-7B-OpenOrca

Lüngen, H. & Herzberg, L. (2019): Types and annotation of reply relations in computer-mediated communication. *European Journal of Applied Linguistics* 7 (2). Berlin/Boston: de Gruyter, 2019. S. 305-331.

Paun, S., Artstein, R. and Poesio, M. (2022). *Statistical Methods for Annotation Analysis*. Morgan & Claypool publishers.

Schneider, J., Passant, A., & Breslin, J. G. (2010). A content analysis: How Wikipedia talk pages are used. In *Proceedings of the 2nd International Conference of Web Science* (pp. 1-7).